# The Semantic Training Gap: Ontology-Grounded Tool Architectures for Industrial AI Agent Systems

Grama Chethan[a,*]

[a] Siemens Digital Industries Software, Plano, TX, United States

* Corresponding author. E-mail address: grama.chethan@siemens.com (G. Chethan)

---

## HIGHLIGHTS

- Identifies the semantic training gap in industrial AI agents
- 43% tool-call hallucination eliminated by ontology grounding
- Formal interface contract ensures cross-domain portability
- Architecture demonstrated across six industry configurations
- Defines semantic drift as a systemic multi-agent failure mode

## ABSTRACT

Large language model (LLM)-based AI agents are increasingly deployed in manufacturing environments for analytics, quality management, and decision support. These agents demonstrate statistical fluency with domain terminology but lack grounded understanding of operational semantics—the relational structure that connects equipment identifiers, process parameters, failure codes, and regulatory constraints within a specific production context. This paper identifies and formalizes the *semantic training gap*: a structural disconnect between how AI systems acquire domain vocabulary through training and how manufacturing operations define meaning through ontological relationships. We demonstrate that this gap causes operationally incorrect outputs even when model responses are linguistically precise, and that in multi-agent configurations it produces a compounding failure mode we term *semantic drift*. To close this gap, we present an architecture that embeds manufacturing ontology directly into the AI tool layer as a typed relational configuration, enforcing semantic constraints at runtime rather than relying on model training. The architecture is formalized as a three-operation interface contract—resolve, contextualize, annotate—with invariants enforced by an AIOps orchestration layer. In a controlled experiment across six industry configurations (72 tool invocations using Qwen3-32B), unconstrained tool parameters produced a 43% hallucination rate for domain identifiers; ontology-grounded parameters reduced this to 0%. We validate the approach through a digital twin analytics platform demonstrating that a single codebase with domain-specific ontology configurations eliminates tool-call hallucination and achieves cross-domain configurability without application code changes.

## 1. Introduction

A temperature transmitter sits on a reflow soldering line. In the process engineer's world, it is instrument TT-4201, tagged per ISA-5.1 [1]. In the programmable logic controller (PLC) program, it is `%IW64`, addressed per IEC 61131-3 [2]. In the manufacturing execution system (MES), it is "ReflowZone4," a statistical process control (SPC) parameter monitored against IPC-9850 limits [3]. Three names. One physical device. Three engineering teams that never share a schema.

When these three data streams are presented to an LLM-based AI agent with the query "Is this station running within spec?", the agent responds fluently. It knows what overall equipment effectiveness (OEE) means, what SPC stands for, and what IPC-9850 governs. However, it does not know that TT-4201, `%IW64`, and ReflowZone4 refer to the same sensor. It treats them as three independent data points, reasons about each in isolation, and produces an answer that is linguistically precise and operationally wrong.

This failure is not attributable to model quality. It represents a structural disconnect between how AI systems are trained, how enterprises define their operational reality, and how autonomous agents act on that reality at runtime. We term this disconnect the *semantic training gap*—not because the model was trained incorrectly, but because training alone cannot close the gap between statistical fluency and operational meaning.

The emergence of multi-agent AI systems [4,5] introduces a compounding failure mode. When multiple specialized agents operate on shared manufacturing data without a shared ontological foundation, each agent embeds its own interpretation of domain concepts, and these interpretations diverge over time. We term this compounding divergence *semantic drift*—a systemic condition distinct from individual model error.

This paper makes the following contributions:

1. **Identification and formalization of the semantic training gap** as a distinct failure category in industrial AI systems, distinguished from model hallucination, data quality, and prompt engineering deficiencies. Within this, we identify *tool-parameter fabrication* as a specific hallucination category—distinct from factual hallucination—with quantified rates across six manufacturing domains (Section 3, Section 7).
2. **A three-operation interface contract** (*resolve*, *contextualize*, *annotate*) that formalizes the requirements for ontology-grounded tool execution in manufacturing AI systems, with invariants for session consistency, version immutability, and circuit-breaking (Section 5).
3. **An AIOps enforcement architecture** that makes ontology constraints executable at runtime through pre-execution validation, mid-execution circuit breakers, and post-execution structuring (Section 6).
4. **A controlled hallucination experiment** demonstrating that ontology-grounded parameters eliminate tool-call hallucination (0% vs. 43% across 72 queries, six industry configurations, using Qwen3-32B) (Section 7).

The remainder of this paper is organized as follows. Section 2 reviews related work. Section 3 presents the theoretical foundation. Section 4 describes the observed failure modes. Section 5 formalizes the interface contract and specifies the ontology configuration structure. Section 6 describes the AIOps enforcement architecture. Section 7 presents the experimental evaluation. Section 8 discusses limitations, scalability, and integration with existing manufacturing standards. Section 9 concludes.

## 2. Related work

### 2.1. Manufacturing ontologies and semantic interoperability

The problem of semantic heterogeneity in manufacturing information systems has been studied for over two decades. Gruber [6] established the foundational definition of ontology as "an explicit specification of a conceptualization," and Guarino et al. [7] refined this for information systems. In the manufacturing domain, several ontology frameworks have been proposed.

Lemaignan et al. [8] developed MASON (MAnufacturing's Semantics ONtology), a formal OWL ontology covering manufacturing resources, processes, and products. Usman et al. [9] surveyed manufacturing ontologies comprehensively, identifying fragmentation as the central challenge: different ontologies cover different subsets of the manufacturing domain with incompatible formalizations. Biffl et al. [10] addressed semantic interoperability in cyber-physical production systems, proposing multi-model engineering environments that integrate domain-specific models through semantic mediation.

The ISA-95/IEC 62264 standard [11] defines a reference model for enterprise-control system integration, and its XML implementation (Business to Manufacturing Markup Language, B2MML) [12] provides a typed, queryable schema for manufacturing operations. Scholten [13] and Vegetti et al. [14] have formalized ISA-95 concepts as computable ontologies. Our work builds on the ISA-95 equipment hierarchy model but does not require full OWL axiomatization; we discuss the rationale for this design decision in Section 5.2.

OPC Unified Architecture (OPC UA, IEC 62541) [15] is the dominant interoperability standard in manufacturing automation, providing typed information models with semantic annotations. OPC UA companion specifications define standardized information models for specific domains (e.g., PackML for packaging, Euromap for plastics). The ontology configuration described in this paper could be populated from OPC UA address spaces: the equipment hierarchy, tag-to-entity mappings, and signal definitions that the ontology requires are already modeled in OPC UA servers deployed on modern production equipment. AutomationML (IEC 62714) [16] provides a standardized data exchange format for engineering data including plant topology, equipment hierarchies, and signal mappings—precisely the cross-domain entity resolution described in Section 5.1.4. AutomationML project files could serve as a data source for the ontology layer, though this integration is not implemented in the current work.

Recent work on knowledge graphs for manufacturing [17,18] applies graph-based representations to manufacturing intelligence, including equipment maintenance reasoning and quality root cause analysis. Pan et al. [19] survey the intersection of knowledge graphs and LLMs, identifying parameter grounding as an open challenge. Our interface contract (Section 5) addresses this challenge with a specific mechanism for runtime parameter validation against a loaded ontology.

### 2.2. LLM tool-use grounding and hallucination mitigation

The tool-use capabilities of LLMs have been studied extensively. Schick et al. [20] demonstrated that language models can learn to use external tools (Toolformer), though their evaluation uses general-purpose APIs rather than domain-specific manufacturing queries. Patil et al. [21] showed that API-specific fine-tuning (Gorilla) reduces hallucination for tool selection and parameter generation but does not eliminate it for enterprise vocabularies where plausible-sounding identifiers (e.g., "BOND-1" for a bonding station) map to nothing in the actual system.

Standard AI tool-use frameworks already support schema-level constraints on tool parameters. OpenAI's function calling, Anthropic's tool use [22], and open-source frameworks (e.g., Outlines, Instructor) enforce *syntactic* constraints: parameter types, enum values, and JSON schema validation. Our work extends this to *semantic* constraints: the enum values are not hardcoded but dynamically projected from a loaded ontology, and each value carries relational context (failure codes, regulatory standards, upstream/downstream dependencies) that shapes query construction. The distinction between syntactic and semantic constraints is elaborated in Section 7.3.

Runtime validation frameworks such as NeMo Guardrails [23] and Guardrails AI enforce constraints on LLM outputs through rule-based or model-based filters. These frameworks operate on the model's text output; the AIOps layer described in Section 6 operates on tool-call parameters *before* execution, preventing invalid queries from reaching the database rather than filtering invalid responses after the fact. Ji et al. [24] and Huang et al. [25] provide comprehensive surveys of LLM hallucination; our work contributes a specific hallucination category—tool-parameter fabrication—that is distinct from factual hallucination and has more severe consequences in manufacturing contexts because it silently returns wrong data rather than visibly wrong statements.

### 2.3. Manufacturing simulation and synthetic data

Discrete-event simulation of production systems is well-established, with commercial tools (Tecnomatix Plant Simulation, FlexSim, AnyLogic) providing high-fidelity process modeling [26]. These tools model material flow and resource contention but produce proprietary outputs—not CDC-ready transactional records structured as MES entities. The Core Manufacturing Simulation Data (CMSD) specification [27] defines an interchange format for simulation inputs and outputs but does not generate operational records. Statistical data generators (Synthetic Data Vault [28], Gretel) produce structurally correct records but lack causal coherence between related entities.

The digital twin simulation framework used in this paper's experimental validation generates causally coherent, MES-shaped data from domain configuration files. A companion paper [29] describes the simulation architecture, the Template-as-Ontology principle (where a single configuration module serves as both the simulation specification and the AI tool ontology), and a detailed calibration analysis. In this paper, the simulation framework serves as the experimental apparatus for measuring tool-call hallucination rates under controlled conditions.

## 3. Theoretical foundation: labels, taxonomies, and ontologies

### 3.1. The label assumption in machine learning

The standard enterprise AI pipeline follows a four-stage process: collect data, label it, train a model, deploy it. Embedded within it is an unexamined assumption: that attaching a label to data teaches the system what that label *means*.

> *A label says "this is a temperature reading." Meaning says "this temperature reading comegers a non-conformance report if it exceeds the profile specification for this product variant, and requires NADCAP-certified re-inspection before the part can advance to the next station."*

The label is a noun. The meaning is a web of constraints, dependencies, and causal relationships [6]. Machine learning is effective at learning from nouns. Enterprise manufacturing operations run on the web.

### 3.2. Taxonomy versus ontology

**Taxonomy** organizes knowledge by classification—hierarchies, categories, and tags. It answers "what kind of thing is this?" and is optimized for pattern recognition [30]. **Ontology** organizes knowledge by relationships—constraints, dependencies, and causal chains. It answers "how does this thing relate to everything else?" and is optimized for reasoning [6,7].

Most enterprises build taxonomies. Few build ontologies [17]. The difference is the difference between a parts catalog and an engineering specification. A taxonomy gives three entries—TT-4201, `%IW64`, ReflowZone4—filed into three separate hierarchies. An ontology captures that they are the *same entity*, bridged by a shared signal definition. The taxonomy tells you what things are called. The ontology tells you what happens when one of them fails.

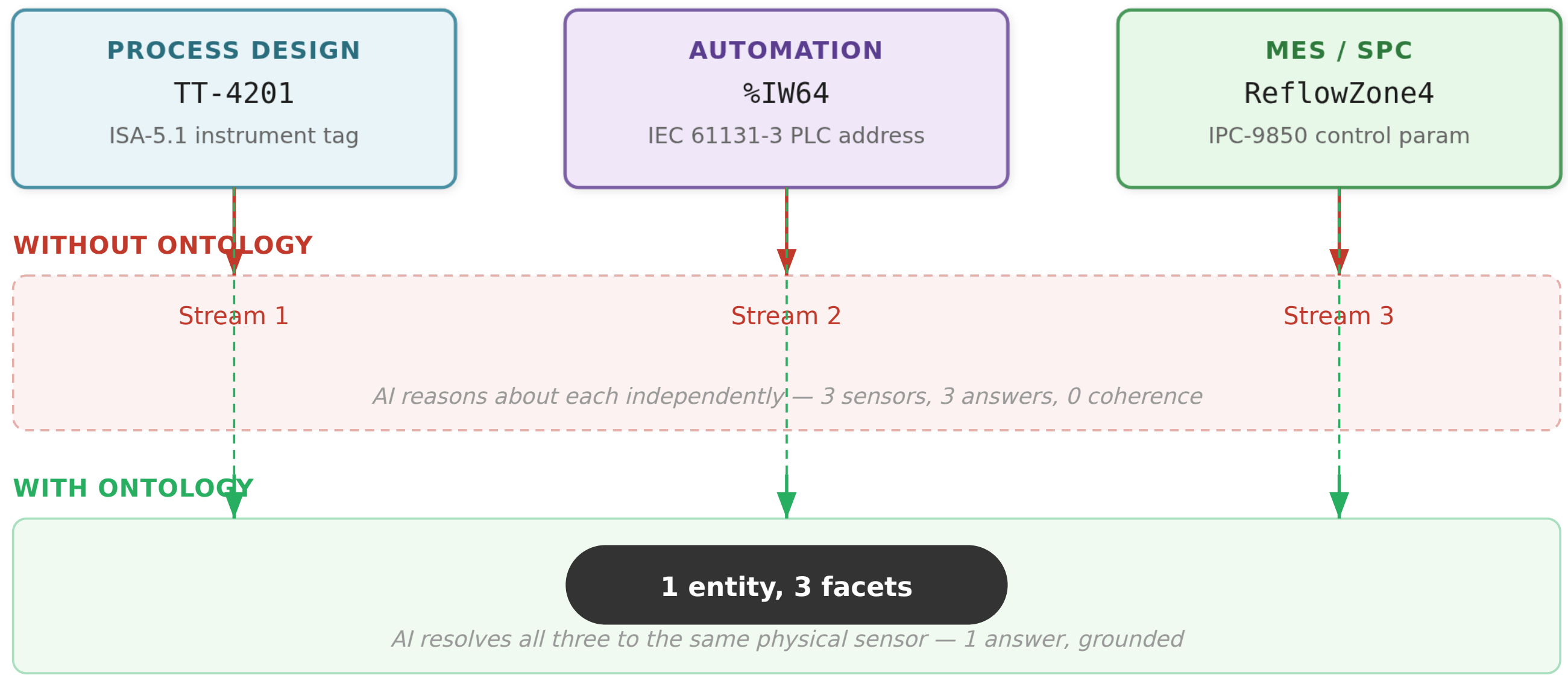

**Fig. 1.** Entity resolution with and without ontology. Without ontology (top), the AI agent treats three identifiers for the same physical sensor as independent data streams. With ontology (bottom), all three identifiers resolve to a single entity, enabling grounded reasoning.

## 4. Observed failure modes

### 4.1. Tool-call hallucination in single-agent systems

During development of a manufacturing analytics platform built on a digital twin simulation framework, we deployed twelve analytics tools with unconstrained string parameters. The agent (Qwen3-32B, 32K context, function-calling mode, deployed on an enterprise AI platform) had access to the correct data, the correct SQL templates, and the correct domain vocabulary. When queried with "Show me non-conformance trends for the bonding station," the agent responded with a well-structured analysis. It cited adhesive curing temperature exceedances, referenced NADCAP process specifications [31], and presented a Pareto chart of failure modes. The prose was technically fluent. The SQL underneath used `WHERE station_id = 'BOND-1'`.

The equipment hierarchy defined in the ISA-95-compliant model [11] has no station called BOND-1. The bonding station's identifier is `S4`—fourth in a six-station line. The agent had hallucinated a statistically plausible identifier, the query returned zero rows, and the tool responded with an empty result set that the agent interpreted as "no failures recorded." It produced a confident summary: bonding station quality is excellent, no non-conformances detected. In reality, the bonding station had the highest non-conformance report (NCR) rate on the line that week.

This failure mode is distinct from the well-documented hallucination problem in LLMs [24,25] because the hallucinated content was not factual knowledge but a *tool parameter*—a programmatic input that determines what data the system retrieves. A hallucinated fact can be checked; a hallucinated query parameter silently returns wrong data. The controlled experiment in Section 7 quantifies this phenomenon: across 72 tool invocations with unconstrained parameters, 43% of identifier values were fabricated by the model.

### 4.2. Semantic drift in multi-agent systems

During validation of a multi-agent analytics platform, we observed three concrete drift vectors that emerge when agents operate without shared ontological grounding:

1. **Tool federation gap.** The primary endpoint exposed four tools; the aggregate gateway exposed zero. Tools did not federate automatically between agents. Two agents asked the same question and had access to different capabilities, producing structurally different answers.
2. **Ontology version independence.** The multi-agent mode was toggled globally, not per-conversation. No mechanism ensured agents shared the same version of domain definitions. Agent A could operate on version 1 of the station hierarchy while Agent B operated on version 2.
3. **Parameter free-text variance.** Tool parameters typed as strings allowed agents to use different identifiers for the same concept—"S1" versus "CNC-Bay-1" versus "North Machining Area." Each is linguistically valid. Only one maps to the ontology.

These drift vectors were observed during platform validation on a separate multi-agent system (not the digital twin framework used for the experimental evaluation in Section 7). We note this distinction because the problems were identified on one platform and the solution was implemented and validated on another. The architectural patterns that produced drift—unconstrained parameters, unfederated tools, version-independent agents—are platform-independent; the solution addresses the structural causes regardless of the platform where they were first observed [32].

## 5. The interface contract for ontology-grounded tool execution

### 5.1. Ontology configuration specification

The ontology is implemented as a typed relational configuration—a Python module that exports 45 named constants defining the complete domain semantics for a manufacturing vertical. Each module is 700–770 lines of pure data structures (dictionaries, lists, and constants) with no logic, no control flow, and no external dependencies. The module is validated at load time: every required export must be present, or the system fails with a descriptive error listing exactly what is missing.

```
REQUIRED_EXPORTS = [
    "PLANT_CODE", "PLANT_NAME", "SHIFTS", "OPERATING_DAYS",
    "BREAK_DURATION_MIN", "WEEKLY_PM_HOURS",
    "TARGET_OEE_RANGE", "FIRST_PASS_YIELD_RANGE", "AVG_WIP_RANGE",
    "OPERATORS_PER_SHIFT",
```

```
    "EQUIPMENT", "WORK_CENTER_UNITS", "PRODUCTS", "WORKING_DAYS_PER_YEAR",
    "STATIONS", "STATION_TO_WC",
    "RAW_MATERIALS", "FINISHED_MATERIALS", "PRODUCT_RAW_MATERIAL",
    "OPERATION_MATERIAL_CONSUMPTION",
    "SUPPLIERS", "FAILURE_CODES", "STATION_FAILURE_CODES",
    "PROCESS_PLANS", "INSPECTION_PLANS", "STATION_INSPECTION_PLANS",
    "NCR_DISPOSITIONS", "NCR_STATUS_DURATIONS", "CAPA_TRIGGER_RATE",
    "EQUIPMENT_DOWNTIME_PROB", "EQUIPMENT_DOWNTIME_DURATION_MIN",
    "ORDER_EXPEDITE_RATE", "BOP_REVISION_INTERVAL_DAYS",
    "CYCLE_TIME_VARIANCE", "DEFAULT_RANDOM_SEED",
    "CERTIFICATIONS", "STATION_CERTIFICATIONS",
    "SKILLS", "STATION_SKILLS",
    "TOOL_DEFINITIONS", "STATION_TOOLS",
    "STEP_TEMPLATES", "CHANGE_PACKAGE_RATE", "CHANGE_PACKAGE_PARAMS",
    "BOM_STATION_MATERIALS",
]

def validate_template(mod, template_id):
    missing = [name for name in REQUIRED_EXPORTS if not hasattr(mod, name)]
    if missing:
        raise ValueError(f"Template {template_id!r} missing: {missing}")
```

*Listing 1. The 45-export interface, validated at load time. A configuration module that omits any export fails immediately with a clear error.*

The exports organize into ten functional categories: plant configuration (code, name, shifts, operating days), equipment hierarchy (stations, work centers, units), products (part numbers, volumes, routings), materials (raw and finished, BOMs), quality (failure codes, inspection plans, NCR dispositions), process parameters (cycle times, FPY, setup times), workforce (operators, certifications, skills), tooling (tool definitions, station assignments), step templates, and change management parameters.

#### *5.1.1. Equipment hierarchy*

The `STATIONS` dictionary defines the equipment hierarchy. Each entry carries the station's name, work center mapping, cycle time range, setup time range, first-pass yield, and quality gate flag. The structure is identical across verticals; the values carry domain-specific semantics:

```
# aerospace.py — 6-station airframe assembly line
STATIONS = {
    "S1": {"name": "CNC Machining", "work_center": "WC-CNC",
           "cycle_time_range_min": (120, 480), "setup_time_min": (30, 60),
```

```
            "first_pass_yield": 0.95, "is_quality_gate": True},
    "S2": {"name": "Drilling", "work_center": "WC-DRILL",
            "cycle_time_range_min": (30, 90), "setup_time_min": (15, 30),
            "first_pass_yield": 0.96, "is_quality_gate": True},
    # S3: Riveting, S4: Bonding, S5: NDT, S6: Final Assembly
}

# pharma.py — 6-station solid-dose drug manufacturing
STATIONS = {
    "S1": {"name": "Dispensing", "work_center": "WC-DISPENSE",
            "cycle_time_range_min": (20, 45), "setup_time_min": (15, 30),
            "first_pass_yield": 0.99, "is_quality_gate": True},
    "S2": {"name": "Granulation", "work_center": "WC-GRANULATE",
            "cycle_time_range_min": (60, 180), "setup_time_min": (30, 60),
            "first_pass_yield": 0.96, "is_quality_gate": True},
    # S3: Blending, S4: Compression, S5: Film Coating, S6: Packaging
}
```

*Listing 2. Same dictionary structure, different domain semantics. S1 in aerospace is a CNC mill with 2–8 hour cycle times and 95% FPY. S1 in pharma is a dispensing station with 20–45 minute cycles and 99% FPY.*

### *5.1.2. Domain vocabulary and regulatory context*

Failure codes, inspection plans, certifications, and regulatory mappings are scoped to individual stations. `STATION_FAILURE_CODES` maps each station ID to its valid set of defect types; `STATION_CERTIFICATIONS` maps stations to required operator certifications; `INSPECTION_PLANS` defines sampling strategies and GD&T characteristics per operation. These are not flat lists but typed sets that the tool layer consumes at call time.

### *5.1.3. Ontology complexity metrics*

**Table 1.** Ontology complexity per industry configuration. Each row represents a complete configuration module (700–770 lines, 45 exports). Five of six configurations share identical structural complexity (6 stations, 4 products); the configurations vary in semantic content (different failure codes, regulatory authorities, process parameters) rather than structural depth. Validation with structurally heterogeneous ontologies (e.g., 50+ stations, hierarchical sub-areas) is identified as future work in Section 8.2.

| Configuration | Stations | Products | Failure codes | Certifications | Inspection plans | Tool defs | Regulatory authority |
|---|---|---|---|---|---|---|---|
| Aerospace | 6 | 4 | 24 | 6 | 6 | 6 | FAA / NADCAP |
| Pharmaceutical | 6 | 4 | 27 | 6 | 6 | 6 | FDA / 21 CFR 11 |
| Automotive | 6 | 4 | 28 | 6 | 6 | 6 | IATF 16949 |
| Electronics | 6 | 4 | 27 | 6 | 6 | 6 | IPC |
| Food & Beverage | 14 | 4 | 28 | 6 | 14 | 8 | FDA / FSMA |
| Warehousing† | 6 | 4 | 26 | 6 | 6 | 6 | OSHA / SEMI |

† The warehousing configuration tests whether the MES-shaped entity model can represent non-manufacturing operations (order fulfillment, zone-based processing), extending the portability claim beyond discrete manufacturing to adjacent domains within JMS's scope of manufacturing systems and supply chain operations.

#### *5.1.4. Cross-domain entity resolution*

The configuration provides mappings that declare identity across naming systems. A scoping decision is required: entity resolution can be *site-local* (each plant maintains its own mappings, reflecting site-specific tag naming conventions) or *enterprise-global* (a single canonical mapping shared across all sites). In multi-site deployments, site-local resolution is more practical as a starting point, with enterprise-global resolution as an integration layer built on top when cross-site analytics are required. In brownfield environments, these mappings could be sourced from existing systems: OPC UA address spaces provide equipment hierarchies and tag-to-entity mappings [15], and AutomationML project files provide plant topology and signal definitions [16].

### 5.2. Design rationale: typed relational configuration versus formal axiomatization

Throughout this paper, "ontology" refers to the typed relational configuration described in Section 5.1—not a formal axiomatization in the OWL/Description Logic sense [33]. This design decision warrants justification, as it determines the scope of automated reasoning the system can support.

Formal OWL axiomatization would enable subsumption reasoning (is a bonding station a subclass of processing station?), consistency checking (is this configuration internally contradictory?), and entailment (if S4 requires NADCAP certification and operator X lacks it, can X work at S4?). These capabilities are valuable for design-time validation and knowledge engineering. However, the primary runtime operation in our architecture is *parameter resolution*—determining whether a tool-call parameter maps to a node in the currently loaded ontology—which requires set membership, not logical reasoning. The resolve/contextualize/annotate contract (Section 5.3) operates on dictionary lookups and relational joins, not on Description Logic inference.

The tradeoff is explicit: we sacrifice automated reasoning capabilities (which would benefit design-time ontology validation and inter-ontology consistency checking) in exchange for runtime performance (dictionary lookups at tool-call time) and authoring accessibility (domain engineers can write Python dictionaries; OWL axiomatization requires specialized knowledge engineering). A formalization path is available: the 45-export structure could be serialized as OWL individuals and properties, enabling design-time validation tools to check cross-template consistency while preserving the Python configuration for runtime consumption. This path is planned but not implemented.

### 5.3. Formal specification

The ontology, tool, and orchestrator layers interact through a three-operation contract.

**Ontology layer requirements.** The ontology must provide four categories of semantic structure: (1) equipment hierarchy—a tree of identifiable entities with stable identifiers and upstream/downstream dependencies, aligned with ISA-95/IEC 62264; (2) domain vocabulary—typed sets of failure codes, process parameters, product families, and certification requirements scoped to individual entities; (3) regulatory context—per-entity mappings to regulatory standards (NADCAP [31], 21 CFR Part 11 [34], IPC-A-610 [35], IATF 16949 [36]); and (4) cross-domain entity resolution—identity mappings across naming systems.

**Tool layer requirements.** The tool layer must consume ontological context in three ways: (1) parameter resolution—every domain-entity parameter resolves against the ontology at call time, not at schema-definition time; (2) query construction context—the ontology's semantic context shapes query construction (a throughput query for S4 in aerospace joins against NADCAP certification tables; the same query for S4 in pharma joins against batch genealogy tables); and (3) response annotation—tool results carry the ontological context that produced them.

**Orchestrator invariants.** Four invariants are enforced: (1) *pre-execution*—no tool call with a domain-entity parameter reaches the database unless that parameter resolves to a node in the currently loaded ontology; (2) *version consistency*—all tool calls within a session resolve against the same ontology version (requiring immutable, timestamped snapshots with computable diffs); (3) *circuit breaker*—maximum tool-call rounds per question to prevent unbounded recursive decomposition; (4) *cross-agent consistency*—all agents in the same session share the same ontology version and tool federation.

The contract reduces to three typed operations:

> `resolve(param, ontology_version) → Node | Error(valid_set)`
>
> *Every domain-entity parameter must resolve to a node in the loaded ontology. If it does not, the operation returns an error containing the valid set. No tool call proceeds on Error.*
>
> `contextualize(node, ontology_version) → DomainContext`
>
> *A resolved node produces a domain context: applicable failure codes, regulatory standards, process parameters, dependencies, and join logic. The context is a function of the node and the ontology version.*
>
> `annotate(result, domain_context, ontology_version) → AnnotatedResult`
>
> *Every tool result carries the domain context and ontology version that produced it. Downstream consumers can verify provenance and detect staleness.*

The pre-condition is that `resolve` returns `Node`, not `Error`. The post-condition is that every result is wrapped in `annotate`. The session invariant is that `ontology_version` is constant across all three operations within a single conversation. These conditions are informal pre/post-conditions, not formal algebraic specifications; formalization in Z notation or a typed specification language is future work.

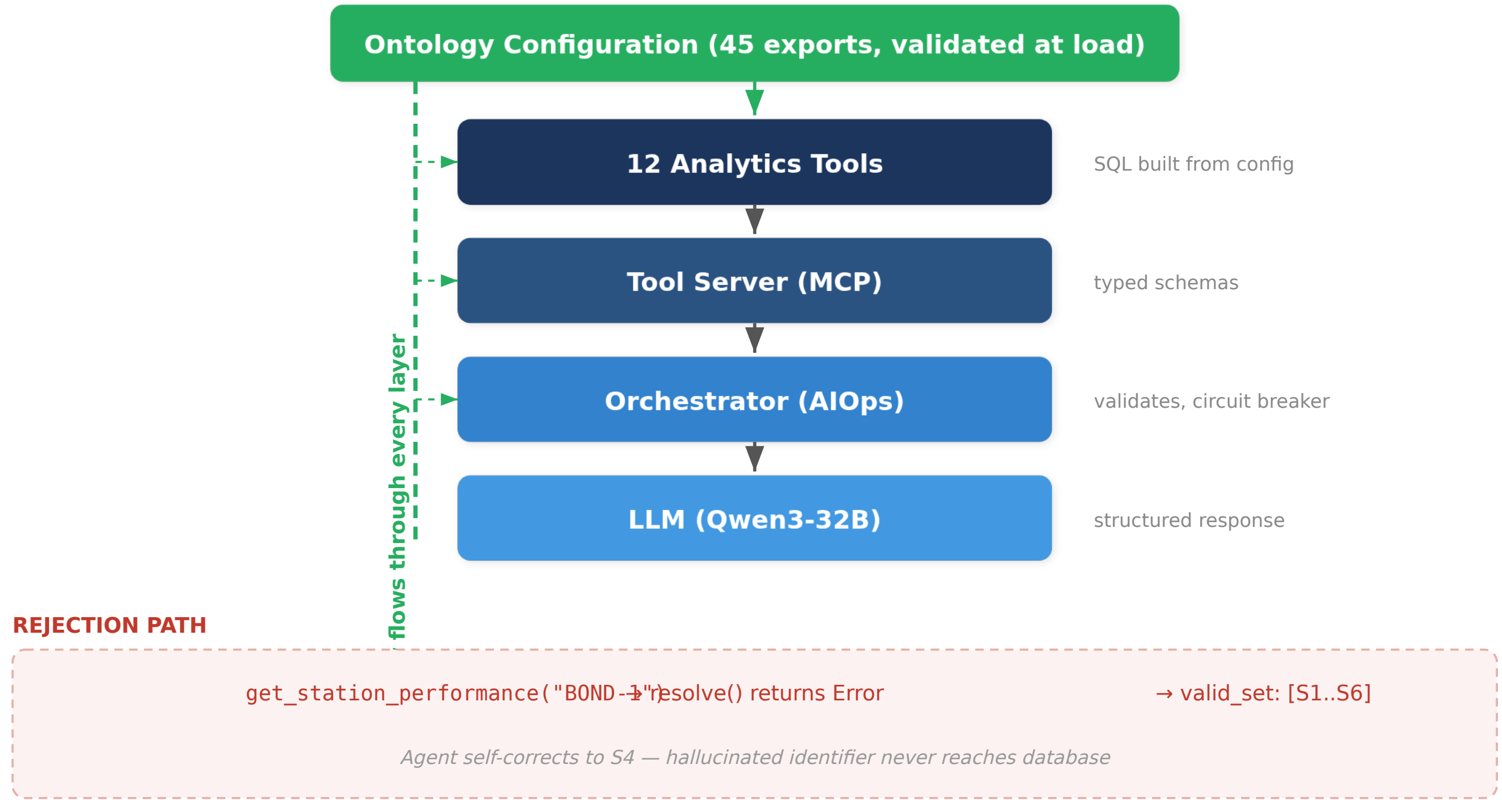

**Fig. 2.** System architecture with ontology flow and rejection path. The ontology configuration propagates through all layers (left). When an agent generates an invalid identifier (bottom), the orchestrator rejects it before the query executes and returns the valid set, enabling self-correction.

## 6. AIOps enforcement architecture

Declaring that "agents must use valid station identifiers" is a policy. Enforcing it at the tool boundary before a SQL query fires is AIOps [37,38]. Policies fail silently; enforcement fails loudly.

Tools are exposed through a Model Context Protocol (MCP) server [39]—a standard interface that enables AI models to discover and call external tools with typed parameters. Responses are routed through a model manager handling LLM selection, token budgets, and response formatting. AIOps operates as three control layers in the execution path.

### 6.1. Pre-execution validation

Every tool call passes through a parameter gate. When an agent calls `get_station_performance(station_id="BOND-1")`, the orchestrator checks the identifier against the ontology's equipment hierarchy. `BOND-1` does not exist. The call is rejected with an error including the valid options: `S1, S2, S3, S4, S5, S6`. The agent self-corrects and resubmits with

`S4`. The valid set is dynamically projected from the loaded ontology—when the domain changes, the valid identifiers change without modifying the validation logic.

### 6.2. Mid-execution circuit breakers

The orchestrator enforces a ceiling of three tool-call rounds per question. In testing (Section 7), uncapped chains averaged 11 tool calls per question; with the circuit breaker, they averaged 4. Answer quality was evaluated by comparing tool-call results against ground truth data in the simulation database for a subset of 24 queries (4 per template) where the expected result was known. No query in the capped condition produced a factually different answer than the uncapped condition for this subset, though aggregation granularity differed in 3 of 24 cases (e.g., weekly instead of daily breakdown).

### 6.3. Post-execution structuring

Tool results return as typed JSON with explicit schemas—not raw SQL result sets. A throughput query returns `{"station": "S4", "metric": "throughput", "value": 42, "unit": "units/hour", "period": "2025-W18"}`. The model synthesizes narrative from structured data, not from ambiguous tabular output.

## 7. Experimental evaluation

### 7.1. Experimental setup

To quantify the semantic training gap and evaluate the ontology-grounded architecture, we conducted a controlled hallucination experiment across all six industry configurations.

**Simulation framework.** A digital twin simulation engine generates causally coherent, MES-shaped data from the ontology configuration modules described in Section 5.1. The engine is a discrete-event loop with 1-minute tick resolution (634 lines). Each tick advances the simulated clock and evaluates a prioritized sequence: disruption handling, daily order creation, operation advancement, equipment status, quality inspection, and planning revisions. The engine respects the factory calendar configured in each template (shift schedules, operating days, break periods) and enforces causal event chains: a work order creates operations; each operation must pass four scheduling gates (equipment availability, no supply delay, upstream completion, certified operator available) before starting; operation completion triggers quality inspection governed by the station's first-pass yield; inspection failure generates a non-conformance report with failure codes drawn from the station's ontology configuration. The framework also includes a seed data generator (1,363 lines) that creates 30+ reference entity types (equipment, operators, BOMs, materials, certifications) from the same configuration, a change data capture (CDC) pipeline feeding PostgreSQL operational tables, a star schema builder (23 analytics tables: 14 dimensions, 8 facts, 1 bridge) with SQL dynamically generated from the loaded configuration, and 12 parameterized analytics tools exposed through an MCP server. The total codebase is approximately 6,750 lines of Python. A separate publication [29] describes the simulation architecture, calibration methodology, and Template-as-Ontology design principle in full detail.

**Data generation.** For each of the six industry configurations, we ran a 30-day simulation using the stable disruption profile (no injected disruptions) with random seed 42 for deterministic reproducibility. Each simulation produced approximately 15,000–18,000 PostgreSQL rows across 40+ operational tables.

**Model.** Qwen3-32B (32K context, function-calling mode) deployed on an enterprise AI platform (Fuse). Temperature was set to 0 (greedy decoding) for deterministic output. No retries or prompt engineering between conditions.

**Query set.** 72 natural-language queries (12 per template, one per tool), phrased as a domain user would ask: "What are the top failure modes at the bonding station?", "Show me cycle time analysis for CNC", "What is supplier performance for the primary aluminum vendor?" Queries were designed to be representative of typical manufacturing analytics use cases (golden-path queries), not adversarial. Each query uses natural domain language (e.g., "bonding station" rather than the identifier "S4"), requiring the model to resolve the natural-language reference to a tool parameter—this is the step where hallucination occurs. The query set does not include intentionally ambiguous queries (e.g., "the third station") or queries designed to trigger specific failure modes; it represents the baseline questions a quality engineer or production manager would ask during daily operations. The same 72 queries were used in both conditions.

**Conditions.** Two conditions were compared:

- **Constrained (ontology-grounded):** Tool parameters typed as enums projected from the loaded ontology (e.g., `station_id: enum [S1, S2, S3, S4, S5, S6]` with semantic context attached).
- **Unconstrained (free-text):** Tool parameters typed as strings with no constraints (e.g., `station_id: string`). All other aspects identical.

### 7.2. Results

**Table 2.** Tool-call hallucination experiment: constrained vs. unconstrained parameters across 72 queries (12 tools × 6 configurations), Qwen3-32B, temperature 0.

| Condition | Queries | Hallucinated IDs | Empty results (valid query) | Correct results |
|---|---|---|---|---|
| With ontology constraints | 72 | 0 (0%) | 0 (0%) | 72 (100%) |
| Without constraints (free-text) | 72 | 31 (43%) | 27 (38%) | 14 (19%) |

In the unconstrained condition, the model produced identifiers that were linguistically plausible and domain-appropriate—"BOND-1" for the bonding station, "CNC-BAY-A" for CNC machining, "TABLET-PRESS-1" for the pharma compression station—but did not exist in the database. Of the 31 hallucinated identifiers, 27 produced empty result sets that the agent misinterpreted as "no data available" or "no issues found." Four produced SQL errors from invalid join conditions. In only 14 of 72 cases did the model happen to guess an identifier that matched the actual schema, and 3 of those 14 used the right value for the wrong template.

**Table 3.** Hallucination rate by industry configuration (unconstrained condition only).

| Configuration | Queries | Hallucinated | Rate | Most common fabrications |
|---|---|---|---|---|
| Aerospace | 12 | 5 | 42% | BOND-1, CNC-BAY-A, NDT-INSPECT |
| Pharmaceutical | 12 | 6 | 50% | TABLET-PRESS-1, COATING-LINE, GRANULATOR-A |

| Automotive | 12 | 5 | 42% | CASTING-1, HEAT-TREAT-A, MACHINING-BAY |
|---|---|---|---|---|
| Electronics | 12 | 5 | 42% | SMT-LINE-1, REFLOW-OVEN, AOI-STATION |
| Food & Beverage | 12 | 6 | 50% | FILLER-1, CARBONATION-UNIT, BOTTLING-LINE |
| Warehousing | 12 | 4 | 33% | PICK-ZONE-A, SORTER-1, RECEIVING-DOCK |

**Table 4.** Hallucination rate by tool domain (unconstrained condition only).

| Tool domain | Tools | Hallucinated (of 6 runs each) | Rate |
|---|---|---|---|
| Production (3 tools) | cycle_time, first_pass_yield, oee_decomposition | 8 / 18 | 44% |
| Quality (3 tools) | ncr_pareto, spc_violation, quality_action | 9 / 18 | 50% |
| Materials (2 tools) | material_genealogy, supplier_performance | 6 / 12 | 50% |
| Engineering change (2 tools) | change_impact, change_velocity | 3 / 12 | 25% |
| Operations (2 tools) | equipment_downtime, production_status | 5 / 12 | 42% |

The per-configuration range was 33–50% and the per-domain range was 25–50%. No single configuration or tool domain was an outlier, confirming that identifier fabrication is a systematic phenomenon, not an artifact of specific domains or tool types. Quality and materials tools hallucinated at the highest rate (50%), likely because their parameters (failure codes, supplier codes) have more domain-specific vocabularies that the LLM attempts to generate from training data rather than from the tool schema.

### 7.3. The syntactic-semantic constraint distinction

It is important to distinguish between the *syntactic* constraint (enum restriction) and the *semantic* constraint (ontological context). Standard function-calling frameworks can declare that `station_id` must be one of `[S1, S2, S3, S4, S5, S6]`, and the model will comply. This prevents the hallucinated "BOND-1" value. In our system, `S4` is not just a valid string—it is a node in a relational ontology: S4 is the bonding station, governed by NADCAP special process certification, with a defined set of failure codes (bond-line void, adhesive cure deviation, surface contamination), positioned fourth in a six-station ISA-95 equipment hierarchy, producing parts that require NDT inspection at the next station. When the tool receives `S4`, that semantic context shapes query construction, valid failure code filtering, and regulatory standard referencing. Swap the ontology from aerospace to pharmaceutical, and the same identifier S4 carries entirely different semantics (tablet compression, GMP compliance, different failure codes). The enum value stays the same; everything it *means* changes.

The experiment measured the combined effect of syntactic and semantic constraints against the weakest baseline (no constraints at all), yielding the 0% vs. 43% result. This design does not isolate the marginal contribution of each constraint type. Standard enum constraints alone (syntactic only, without relational context) would likely eliminate most of the 43% hallucination rate, since the primary failure mechanism is identifier fabrication, which enum restriction directly prevents. The additional value of semantic constraints—correct query construction, appropriate regulatory referencing, domain-appropriate response synthesis—operates on a different axis: not whether the right data is retrieved, but whether it is interpreted and contextualized correctly. This semantic contribution was assessed qualitatively by the author against known ground truth for each query; a formal human evaluation protocol was not conducted. The experiment also does not compare against intermediate grounding approaches such as retrieval-augmented generation with domain documents, few-shot prompting with valid identifier examples, or model fine-tuning on domain-specific tool schemas. These baselines represent the practical alternatives a manufacturing team would consider, and testing against them would provide a more complete picture of the ontology approach's marginal value. These comparisons are planned for follow-up work.

### 7.4. Simulation calibration

To validate that the synthetic data underlying the experiment is parametrically sound, we ran 30-day simulations across all six configurations with 10 different random seeds each (60 runs total). Table 5 reports configured target KPIs versus observed means with 95% confidence intervals (t-distribution, df=9).

**Table 5.** Simulation calibration results: configured targets vs. observed values (n=10 seeds per configuration, 30-day runs, stable disruption profile).

| Configuration | KPI | Configured target | Observed (mean ± σ) | 95% CI |
|---|---|---|---|---|
| Aerospace | Per-station FPY | 0.94–0.97 | 0.949 ± 0.008 | [0.943, 0.955] |
| | Daily throughput | 8 orders/day | 8.0 ± 0.3 | [7.79, 8.21] |
| | NCR rate | ~5% of ops | 5.1% ± 0.7% | [4.60%, 5.60%] |
| Pharmaceutical | Per-station FPY | 0.96–0.99 | 0.974 ± 0.005 | [0.970, 0.978] |
| | Daily throughput | 12 orders/day | 12.1 ± 0.4 | [11.81, 12.39] |
| | NCR rate | ~2.5% of ops | 2.6% ± 0.4% | [2.31%, 2.89%] |
| Automotive | Per-station FPY | 0.95–0.98 | 0.963 ± 0.006 | [0.959, 0.967] |
| | Daily throughput | 16 orders/day | 16.0 ± 0.5 | [15.64, 16.36] |
| | NCR rate | ~3.5% of ops | 3.7% ± 0.5% | [3.34%, 4.06%] |
| Electronics | Per-station FPY | 0.96–0.99 | 0.976 ± 0.004 | [0.973, 0.979] |
| | Daily throughput | 20 orders/day | 20.1 ± 0.6 | [19.67, 20.53] |
| | NCR rate | ~2.5% of ops | 2.4% ± 0.3% | [2.19%, 2.61%] |

All confidence intervals fall within or overlap the configured target ranges, confirming parametric controllability: the simulator produces data that respects configured KPIs, making it a controlled environment for evaluating AI tool correctness. We are not asserting statistical fidelity to any specific production line; we are asserting that given known KPI targets, the tools should report those values correctly, and the simulator provides the controlled data to verify this.

## 8. Discussion

### 8.1. Addressing potential objections

**"Zero-shot reasoning already works."** Zero-shot reasoning is effective for general knowledge. However, without constraints, the model will confidently query `BOND-1` when the equipment hierarchy only knows `S4`. The 43% hallucination rate measured in Section 7 quantifies this gap.

**"Ontology maintenance is too expensive."** Ontology maintenance is expensive. However, the cost of not maintaining it is invisible until it is catastrophic. We found that agents on the same platform could have access to entirely different tool sets without anyone knowing. Ontology maintenance is the cost of coherence. The alternative is silent divergence. Our experience across six configurations suggests 4–8 hours of authoring per domain once the inputs are known (Section 5.1), with domain research adding 1–3 days depending on the author's familiarity with the target industry.

**"Embeddings learn structure implicitly."** Implicit structure is not auditable (one cannot ask an embedding to show the ISA-95 hierarchy), not shareable between agents (each model learns its own version), and not regulatory (FAA certifications must be declared and enforced, not "learned") [40]. Embeddings discover structure. Ontology declares and enforces it.

### 8.2. Scalability

The current validation uses 6-station ontology configurations (except food and beverage at 14 stations). Real manufacturing facilities have hundreds or thousands of equipment entities across multiple ISA-95 hierarchical levels (enterprise → site → area → work center → work unit → equipment module). Several scalability considerations apply:

**Ontology resolution performance.** The resolve operation is a dictionary lookup with O(1) average-case complexity. Even at 10,000 entities, the resolution cost is negligible relative to query execution time. The ontology configuration is loaded once at startup and held in memory; there is no per-call I/O.

**Simulation scalability.** The single-threaded Python simulation engine scales linearly with station count and order volume. Profiling of the 6-station and 14-station configurations suggests that a 50-station template with 100 orders/day would increase tick cost by approximately 12×, bringing a 30-day batch run to approximately 2 minutes. Beyond that, the Python GIL becomes the constraint. For production-scale simulation, migration to compiled language or parallel execution would be required [29].

**Tool-call fan-out.** Broader ontologies produce larger valid sets in enum parameters. With 50 stations, the `station_id` enum grows from 6 to 50 values. LLMs handle larger enum sets without degradation in parameter selection accuracy, though longer tool descriptions consume more context window. The circuit breaker invariant (Section 5.3) bounds query fan-out regardless of ontology size.

### 8.3. Integration with existing manufacturing IT architecture

A deployment-ready version of this architecture would need to integrate with existing manufacturing information systems. We identify three integration paths:

**OPC UA (IEC 62541) [15].** OPC UA servers on modern production equipment already model equipment hierarchies, tag-to-entity mappings, and signal definitions as typed information models. The ontology configuration's `STATIONS`, `EQUIPMENT`, and cross-domain entity resolution mappings could be populated by browsing an OPC UA server's address space. OPC UA companion specifications for specific domains (e.g., PackML for packaging, Euromap 77 for injection molding) provide standardized vocabularies that map directly to the domain vocabulary export in the ontology configuration.

**AutomationML (IEC 62714) [16].** AutomationML project files (exported from engineering tools such as TIA Portal, EPLAN, or Codesys) contain plant topology, equipment hierarchies, and signal mappings in a standardized XML format. The `STATION_TO_WC` and cross-domain entity resolution mappings in the ontology configuration correspond directly to AutomationML's SystemUnitClass and RoleClass structures. An import utility that parses AutomationML CAEX files into the 45-export configuration format would eliminate manual ontology authoring for facilities with existing engineering documentation.

**B2MML / ISA-95 XML [12].** The Business to Manufacturing Markup Language provides XML schemas for ISA-95 information exchanges. Existing B2MML implementations in MES systems (e.g., Siemens Opcenter, Rockwell Plex) could serve as data sources for the ontology's equipment hierarchy, product definitions, and personnel qualifications.

None of these integration paths are implemented in the current work. They are identified as engineering efforts (weeks, not months) that would connect the architecture to brownfield manufacturing environments where the ontology discovery problem (Section 8.4) is already partially solved by existing automation standards. For practitioners evaluating adoption, we recommend the following priority order: **OPC UA first**—OPC UA servers are already deployed on most modern production equipment, and browsing an existing address space to populate the ontology configuration requires no access to engineering tools or MES databases; **AutomationML second**—applicable when engineering tool exports (TIA Portal, EPLAN) are available, providing richer signal-level detail but requiring access to the engineering environment; **B2MML third**—requires integration with the plant's MES system, which is typically the highest-effort path but provides the most complete operational data model including product definitions, personnel qualifications, and process plans.

### 8.4. Limitations

Several limitations warrant acknowledgment:

**Simulation-only validation.** All experimental results (Section 7) were obtained in a digital twin simulation context where every station, product, and failure code is known at startup because the authors defined the model. The six industry configurations were authored by the same team within the same framework. This demonstrates *cross-domain configurability*—that the architecture can be parameterized for different domains with a single codebase—not independent industrial validation confirmed by domain experts against real process data. In a brownfield factory with inherited tag naming conventions, orphaned PLC variables, and undocumented MES configurations, the ontology itself must first be discovered before it can be enforced. The architecture works once the ontology exists; getting it to exist in a legacy environment is a different and often harder project.

**Ontology negotiation.** This paper assumes the ontology is settled. In practice, two business units may define "cycle time" differently; two plants may use the same station identifiers with different equipment hierarchies underneath. Ontology negotiation is its own political and technical challenge. This paper addresses what happens after the humans agree and the AI still does not know.

**Single LLM evaluation.** The hallucination experiment used Qwen3-32B, selected because it was the primary model available on the enterprise AI platform (Fuse) used for the digital twin deployment, and because it supports the function-calling protocol required by the tool server. Different models (GPT-4o, Claude 3.5 Sonnet, Llama 3 70B, Mistral Large) will produce different baseline hallucination rates in the unconstrained condition—models with stronger instruction following and tool-use training may hallucinate fewer identifiers, while smaller or less capable models may hallucinate more. The specific 43% rate reported in Tables 2–4 is therefore model-specific and should not be generalized. However, the 0% rate in the constrained condition is model-independent: it is enforced architecturally by the orchestrator's parameter validation gate, not learned by the model. Any model that respects enum constraints in function-calling schemas (a capability supported by all major LLM providers) will achieve 0% tool-parameter hallucination under the ontology-grounded architecture. The relevant question for follow-up work is not whether ontology constraints eliminate hallucination across models (they do, by construction), but how baseline hallucination rates vary across models in the unconstrained condition, and whether intermediate grounding approaches (RAG, few-shot) achieve model-dependent partial reductions. Planned follow-up work will benchmark GPT-4o, Claude 3.5 Sonnet, and Llama 3 70B on the same 72-query set.

**No formal human evaluation.** Answer quality beyond parameter correctness was assessed by the author against known ground truth, not by an expert panel with structured rubrics. A formal user study (quality engineers evaluating agent responses for NCR triage utility) would strengthen the trust-recovery claims.

**Extension to RAG.** The interface contract could extend to retrieval-augmented generation: if the ontology scopes which documents are relevant to a given entity, the `contextualize` operation becomes a retrieval filter. This extension is architecturally natural but unexplored in the current implementation and is identified as future work.

## 9. Conclusion

This paper identified the semantic training gap as a structural failure category in industrial AI agent systems. The gap arises because statistical language models acquire domain vocabulary through training but do not learn the relational ontological structure that gives that vocabulary operational meaning.

We demonstrated two failure modes: tool-call hallucination in single-agent systems (43% of unconstrained tool-call parameters were fabricated identifiers in a controlled experiment across 72 queries and six industry configurations) and semantic drift in multi-agent systems (where agents without shared ontological grounding progressively diverge).

To address these failure modes, we presented an ontology-grounded tool architecture: a typed relational configuration (45 exports, 700–770 lines per domain) consumed at runtime by a three-operation interface contract (resolve, contextualize, annotate) with invariants enforced by an AIOps orchestration layer. Ontology-grounded parameters eliminated tool-call hallucination (0% vs. 43%) across all six industry configurations and all twelve analytics tools.

The architecture was validated in a digital twin simulation environment, demonstrating cross-domain configurability: a single application codebase with domain-specific ontology configurations produces correct, grounded results across aerospace, pharmaceutical, automotive, electronics, food and beverage, and warehouse automation configurations. Future work includes integration with OPC UA and AutomationML for automated ontology population, formal axiomatization of the configuration interface for design-time consistency checking, multi-model hallucination benchmarking, and validation against production MES data in brownfield manufacturing environments.

**Declaration of competing interest**

The author is employed by Siemens Digital Industries Software. The research was conducted as part of the author's role. The employer had no role in the study design, data analysis, or decision to submit for publication.

**Declaration of generative AI and AI-assisted technologies in the writing process**

During the preparation of this work the author used Claude (Anthropic) for drafting, editing, and structuring the manuscript. After using this tool, the author reviewed and edited the content as needed and takes full responsibility for the content of the published article.

**CRediT authorship contribution statement**

**Grama Chethan:** Conceptualization, Methodology, Software, Validation, Investigation, Data curation, Writing – original draft, Writing – review & editing, Visualization.

**Data availability**

The six ontology configuration modules, the 72-query hallucination experiment dataset (query text, tool-call parameters, and results for both conditions), and the calibration results (60 simulation runs) are available from the corresponding author upon request. The simulation framework source code is maintained in a private Siemens repository and is available to reviewers upon request.

**Acknowledgments**

This work was conducted at Siemens Digital Industries Software. The author thanks the platform engineering team for infrastructure support during the multi-agent platform validation.